\documentclass[journal=jacsat,manuscript=article]{achemso}
\usepackage[version=3]{mhchem} 

\usepackage[utf8]{inputenc}
\usepackage[T1]{fontenc}    
\usepackage{hyperref}       
\usepackage{url}            
\usepackage{booktabs}       
\usepackage{wrapfig,lipsum}
\usepackage{amssymb}
\usepackage{amsfonts}       
\usepackage{nicefrac}       
\usepackage{microtype}      
\usepackage{xcolor}         
\usepackage{graphicx}
\usepackage{amsmath}
\usepackage[normalem]{ulem}
\useunder{\uline}{\ul}{}

\author{Àlex Solé}
\affiliation[Universitat Politècnica de Catalunya]
{Image Processing Group - Signal Theory and Communications Department, Universitat Politècnica de Catalunya, Barcelona, Spain}
\alsoaffiliation[Universitat de Barcelona]
{Inorganic and Organic Chemistry Department and Institute of Theoretical and Computational Chemistry, Universitat de Barcelona, Barcelona, Spain}

\author{Albert Mosella-Montoro}
\affiliation[Universitat Politècnica de Catalunya]
{Image Processing Group - Signal Theory and Communications Department, Universitat Politècnica de Catalunya, Barcelona, Spain}

\author{Joan Cardona}
\affiliation[Universitat de Barcelona]
{Inorganic and Organic Chemistry Department and Institute of Theoretical and Computational Chemistry, Universitat de Barcelona, Barcelona, Spain}

\author{Daniel Aravena}
\email{daniel.aravena.p@usach.com}
\affiliation[Universidad de Santiago de Chile]
{Materials Chemistry Department, Faculty of Chemistry and Biology, Universidad de Santiago de Chile, Santiago, Chile}

\author{Silvia Gómez-Coca}
\email{silvia.gomez@qi.ub.es}
\affiliation[Universitat de Barcelona]
{Inorganic and Organic Chemistry Department and Institute of Theoretical and Computational Chemistry, Universitat de Barcelona, Barcelona, Spain}

\author{Eliseo Ruiz}
\email{eliseo.ruiz@qi.ub.edu}
\affiliation[Universitat de Barcelona]
{Inorganic and Organic Chemistry Department and Institute of Theoretical and Computational Chemistry, Universitat de Barcelona, Barcelona, Spain}

\author{Javier Ruiz-Hidalgo}
\email{j.ruiz@upc.edu}
\affiliation[Universitat Politècnica de Catalunya]
{Image Processing Group - Signal Theory and Communications Department, Universitat Politècnica de Catalunya, Barcelona, Spain}

\title{PRISM: Periodic Representation with multIscale and Similarity graph Modelling for enhanced crystal structure property prediction}

\begin{document}

\begin{abstract}
Crystal structures are characterised by repeating atomic patterns within unit cells across three-dimensional space, posing unique challenges for graph-based representation learning. Current methods often overlook essential periodic boundary conditions and multiscale interactions inherent to crystalline structures. In this paper, we introduce PRISM, a graph neural network framework that explicitly integrates multiscale representations and periodic feature encoding by employing a set of expert modules, each specialised in encoding distinct structural and chemical aspects of periodic systems. Extensive experiments across crystal structure-based benchmarks demonstrate that PRISM improves state-of-the-art predictive accuracy, significantly enhancing crystal property prediction. The code will be released on GitHub upon acceptance.
\end{abstract}

\section{Introduction}

Accurate prediction of crystal material properties is essential for accelerating the discovery of novel materials for applications such as energy storage, catalysis, and electronics~\cite{solar_cells, batteries,mat_appl, superconductor, batteries2, catalysis}. While density functional theory (DFT) provides reliable predictions~\cite{DFT_reliability}, its high computational cost limits its applicability for large-scale screening~\cite{DFT_comp_cost}. Consequently, machine learning approaches that leverage graph representations of crystal structures have emerged as a promising alternative~\cite{CGCNN,megnet,CATGNN,alignn}.

Recent works \cite{conformer, CARTNET} use graph neural networks where the graph is created using atom geometric proximity policies such as: k-Nearest Neighbours (k-NN) or radius. However, these techniques do not explicitly encode the unit cell; they focus solely on atom-level interactions with periodic boundary conditions applied in the geometrical space. Moreover, those techniques struggle to capture long-distance interactions, which are crucial for understanding the global structural context of crystalline materials.

Furthermore, a common strategy in molecules or 3D point clouds  tasks~\cite{MOSELLAMONTORO2021,frank2022so3krates,SPOTR,PointConT} is to employ a double-neighbourhood approach that captures information from both the geometrical domain and the feature space.  However, when applied to crystals, it is crucial to encode periodicity directly within the feature space. Without this encoding, the feature space remains oblivious to the inherent repeating patterns, which can ultimately limit predictive performance. Additionally, as demonstrated by Keqiang et al.~\cite{conformer}, the same material can have multiple equivalent cell representations. This variability must be addressed during the feature graph construction to ensure invariance with respect to the cell representation.

To address these limitations, we propose an extended framework that explicitly incorporates unit cell information and enforces the periodicity in the feature spaces. Our contributions are as follows:
\begin{itemize}

  \item \textbf{Cell-Level Structural Encoding:} We introduce a dedicated module that captures and encodes cell-level structural details by adding a cell node, providing a comprehensive global perspective of the crystal's periodic architecture.
  
  \item \textbf{Periodic Feature Encoding:} We develop a specialised module that explicitly incorporates periodic boundary conditions into the feature space, ensuring that the inherent repeating patterns and symmetry of crystalline materials are fully captured, ensuring invariance with respect to the cell representation.
  
  \item \textbf{Multiscale Fusion:} We design a comprehensive fusion module that seamlessly integrates the cell-level details with atom-level representations into a cohesive framework, thereby enhancing the overall predictive performance of the network.

  \item \textbf{Mixture-of-Experts Integration:} We introduce a principled mixture-of-experts mechanism that jointly leverages the cell-level structural encoding, periodic feature encoding, and multiscale, learning property-dependent gates and fusion weights to combine the different experts. This integration delivers strong predictive performance while remaining interpretable, by explicitly showing how much each expert contributes to each prediction.
\end{itemize}

We validate our approach by benchmarking it on Jarvis~\cite{Jarvis}, Materials Project~\cite{megnet}, and Matbench~\cite{dunn2020benchmarking} datasets. Our extensive experiments demonstrate that by explicitly encoding the unit cell and enforcing the periodicity in the feature space, our framework significantly improves predictive accuracy, paving the way for more accurate models in crystal property prediction. The code will be released on GitHub upon acceptance.

\section{Related Work}

\paragraph{Crystal GNNs.} Graph Neural Networks (GNNs) have served as a powerful framework for modeling periodic crystal structures by representing atoms as nodes and interatomic relationships as edges. Early methods such as CGCNN~\cite{CGCNN} introduced multi-edge crystal graphs built upon Euclidean distances to capture short-range atomic interactions under periodic boundary conditions. Extensions like MEGNET~\cite{megnet} adopted enriched atom and bond embeddings, while GATGNN~\cite{CATGNN} incorporated attention mechanisms to aggregate local chemical and structural features more effectively. Angle-based line graphs in ALIGNN~\cite{alignn} and M3GNet~\cite{chen2022graph} further enhance geometric expressiveness, although at an additional computational cost tied to the average number of neighbours per atom. Techniques like MatFormer~\cite{matformer} use self-loops for encoding periodic pattern awareness, whereas PotNet~\cite{potnet} approximates infinite interactions between every pair of atoms, incurring a quadratic overhead in the number of atoms. More recent work (eComformer~\cite{conformer}, iComformer~\cite{conformer}, and CartNet~\cite{CARTNET}) address orientation ambiguities, either by constructing orientation-invariant or -equivariant representations of the lattice vectors, or by combining rotational data augmentation with explicit direction vectors. Nonetheless, most rely primarily on distance-only or partially invariant embeddings, which can limit the capacity to model certain anisotropic or long-distance crystal properties.

\paragraph{Feature-Space Graphs.}
Beyond geometry-based neighbourhoods such as fixed k-NN or radius graphs~\cite{Mosella2019RAGC, pointnet++}, connecting points directly in the learned feature space often yields richer task-specific relations. DGCNN~\cite{DGCNN} recomputes the k-NN graph from the current feature space at every layer so edges link feature-nearest rather than Euclidean-nearest points and semantic information can move between regions that are far apart in space. Building on this principle, MUNEGC~\cite{MOSELLAMONTORO2021} constructs one graph in Euclidean space to preserve local geometric context and another in feature space to connect semantically similar yet spatially distant points. It then aggregates messages from both graphs to obtain a representation that blends geometric and semantic cues. 
In molecular modelling, So3krates~\cite{frank2022so3krates} further extends these ideas by introducing feature-based neighbours that capture nonlocal quantum interactions, moving beyond traditional short-range or coordinate-based connections. 
Nevertheless, these methods were invitally developed for finite or nonrepetitive 3D tasks and often omit periodic boundary conditions, which complicates their application to crystalline systems. Adapting these ideas to periodic crystals is vital, since each feature neighbourhood must accurately reflect a unique representation of the crystal space, independently of the multiple equivalent cell representations.

Despite their popularity, fixed‐radius or k-NN graphs limit message passing to local neighbourhoods and cannot capture the long‐distance lattice correlations that often govern crystal properties. Likewise, feature-space graph methods, developed primarily for finite point clouds, omit explicit periodic boundary conditions, thus failing to exploit translational symmetry. Our approach addresses both shortcomings by enforcing minimal‐image periodicity in its feature‐space graph construction and fusing this with multiscale atomistic and cell‐level graphs, enabling unified encoding of short‐range chemistry and long‐range structural interactions.  

\section{Methodology}

PRISM (Periodic Representation with multIscale and Similarity graph Modelling) is a multigraph-based neural network architecture tailored for crystalline materials. It employs a collection of expert modules, each specialised in encoding complementary structural and chemical features of periodic systems. Drawing inspiration from ensemble learning~\cite{ensemble_review,ensemble_review_2}, every expert operates on a distinct graph topology to capture interactions at different spatial scales, thereby enabling PRISM to jointly model both local atomic environments and global lattice periodicity.

\begin{figure}[h]
    \centering
    \includegraphics[width=\linewidth]{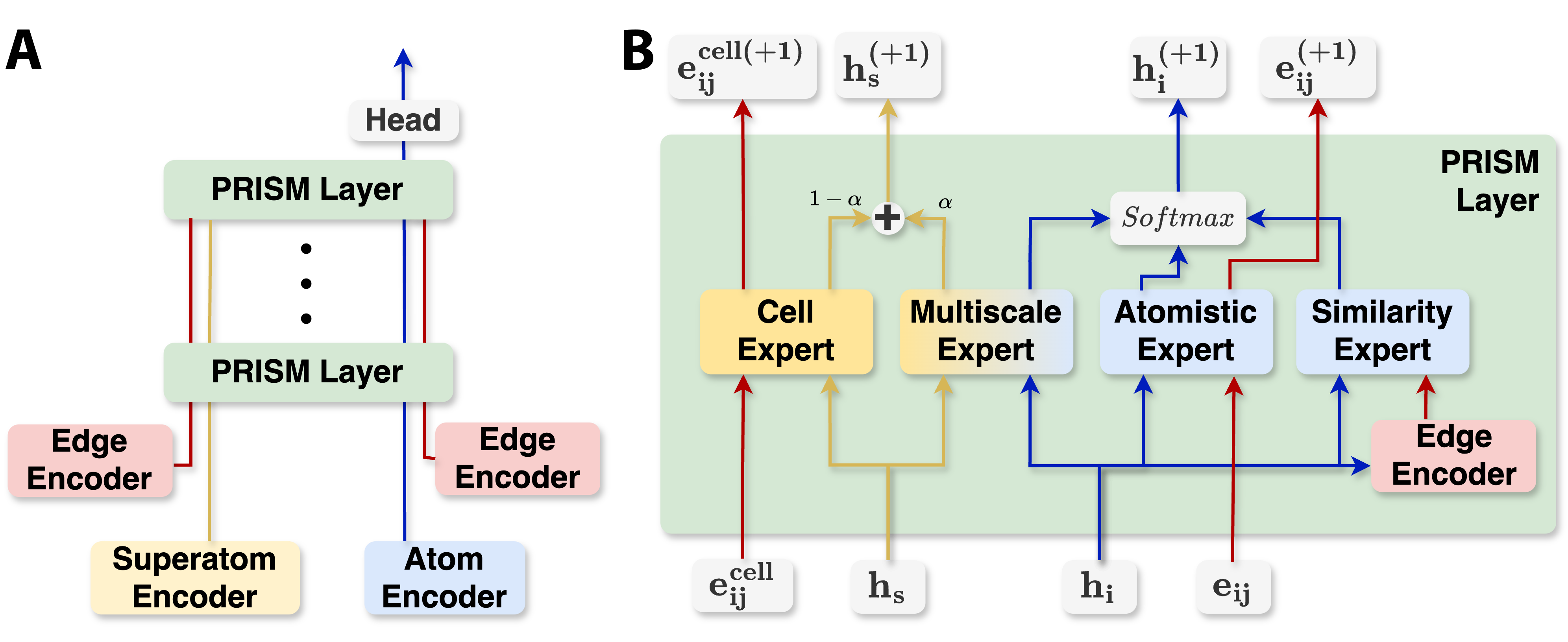}
    \caption{
    \textbf{A} Overview of the PRISM architecture. Atomic and superatom embeddings are initialised via dedicated encoders, and each PRISM layer updates them until the final representation that is used to predict the property. \textbf{B} The PRISM layer architecture. Each of the representations are aggregated and fused using four expert modules: Cell (global lattice periodicity), Multiscale (atom-superatom bidirectional interactions), Atomistic (radius graph with periodic boundaries), and Similarity (feature-space graph with periodic boundaries). 
    }
    \label{fig:baseline}
\end{figure}

Our PRISM framework constructs a dual-scale, multi-graph representation that is iteratively refined through an ensemble of experts. As illustrated in Figure~\ref{fig:baseline}a, we maintain both atomistic embeddings $\mathbf{h}_i$ (highlighted in blue) and a global superatom embedding $\mathbf{h}_s$ (highlighted in yellow), each connected via their respective edge sets. Each PRISM layer aggregates and fuses information through four specialised experts: the \emph{Cell Expert} encodes lattice periodicity by constructing a superatom graph that captures global cell-level repetitions; the \emph{Multiscale Expert} mediates bidirectional information flow between atomistic and superatom embeddings to integrate local and global context; the \emph{Atomistic Expert} models short-range interatomic interactions through a radial cutoff graph under periodic boundary conditions; and the \emph{Similarity Expert} connects atoms with similar electronic environments in feature space under periodic boundary conditions, thereby enabling long-range relational modeling while preserving periodic invariance. All embeddings are updated in parallel within each layer, and the final atomistic embeddings are used to predict the target property (see Figure~\ref{fig:baseline}B).             

To formalise the input to our PRISM architecture, we represent each crystal as a graph over its atomic sites and periodic lattice vectors, defined as follows. Given a crystal structure, we define the set of atoms, represented as nodes $\mathcal{V}$, which represents the $N$ atoms within a unit cell. Each node $i \in \mathcal{V}$ is characterised by its Cartesian coordinates $\mathbf{r}_i \in \mathbb{R}^{3}$ and an initial latent feature embedding $\mathbf{h}_i^{(0)} \in \mathbb{R}^{dim}$ derived from its atomic number, where $dim$ denotes the dimensionality of the latent feature vector. The periodic boundaries of the crystal are defined by the lattice matrix $\mathbf{L} = [\mathbf{l}_1, \mathbf{l}_2, \mathbf{l}_3] \in \mathbb{R}^{3\times3}$, with lattice vectors $\mathbf{l}_j \in \mathbb{R}^{3}$ specifying the geometry of the unit cell. Based on this formalism, we propose four distinct expert modules, each designed to encode the crystal's complementary structural and chemical characteristics by operating over specialised graph topologies defined on the atomic set $\mathcal{V}$.

\subsection{Atomistic Expert}

To accurately model the fundamental chemical bonds and local physical forces that govern crystalline stability and emergent properties, it is essential to capture interactions at the atomistic scale. This expert captures short‐range atomic interactions by constructing a graph $\mathcal{G}_{\mathrm{atomistic}}$ connecting atoms $i$ and $j$ when their minimum‐image distance under Periodic Boundary Conditions satisfies:
\[
\|\mathbf{r}_i - \mathbf{r}_j + \mathbf{L}\mathbf{n}_{lmn}\| < r_c,
\]
where $\mathbf{n}_{lmn}\in\mathbb{Z}^3$ is the lattice shift vector of the image.

\begin{figure}[h]
    \centering
    \includegraphics[width=\linewidth]{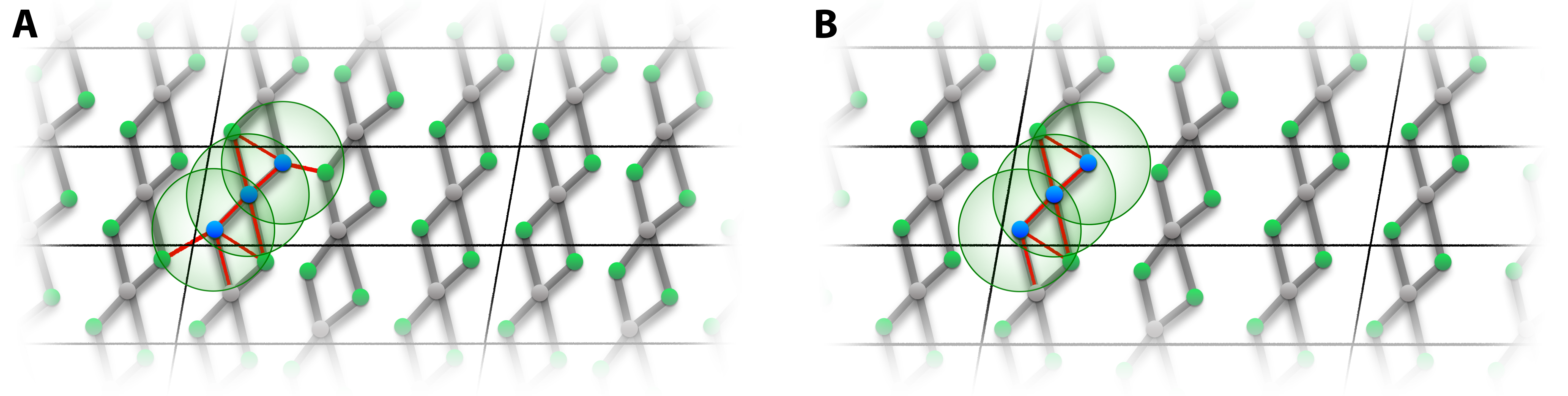}
    \caption{
    Atomistic expert graph $\mathcal{G}_{\mathrm{atomistic}}$ under periodic boundary conditions. A radius-based neighbourhood with cutoff $r_c$ (green spheres) is built around each reference atom (blue), and edges (red) connect atoms within $r_c$ while accounting for periodic images. \textbf{A} Three reference atoms are highlighted, and a radius graph is created around each of them. The resulting lateral links connect adjacent layers of the material, allowing message passing to propagate across layers and correctly encode the geometry. \textbf{B} In this material, the cutoff does not reach lateral neighbours, so edges form mainly along the vertical direction. Message passing is then propagated only vertically, and the model cannot capture the lateral geometry; stacking additional message-passing layers does not restore the missing lateral pathways.
    }
    \label{fig:pbc_radius_graph}
\end{figure}

Although a radius-based construction is one of the most common choices when building graphs for GNNs for materials, it has an important limitation about propagating the information, since it assumes that all the regions are connected. If two regions are not connected through the graph, messages cannot propagate across these gaps, and even stacking multiple message-passing layers that follow this connectivity does not restore the missing pathways. Increasing $r_c$ can mitigate this by adding longer-range edges. This, however, weakens locality, often degrades overall performance, and has an expensive computational cost, as discussed in the Discussion Section. As illustrated in Figure~\ref{fig:pbc_radius_graph}, when the layers of the material are close enough to have a connection inside the $r_c$, the information is propagated through all the system. When the material layers are more separated, the information is only propagated vertically, causing that the message passing is not encoding the structure correctly.

\subsection{Similarity Expert}
\label{sec:similarity}

To address the propagation bottlenecks that the atomistic radius graph $\mathcal{G}_{\mathrm{atomistic}}$ can induce, we introduce the \textit{Similarity} expert. This expert propagates semantically relevant chemical and structural cues across distant yet similar atoms, capturing global correlations that spatial proximity alone cannot reveal. It explicitly models similarity by constructing a feature-space graph, $\mathcal{G}_{\mathrm{feat}}$, where atoms are connected if the Euclidean distance between their learned feature embeddings is below a feature cutoff $r_f$. Because our message-passing modules encode geometry on edges, we then instantiate geometric edges between these feature-connected nodes using their minimum-periodic displacement. Linking atoms with similar chemistry or local environments enables global information flow and supports long-range dependencies beyond radius-only neighbourhoods.

For each edge connecting two atoms $i$ and $j$ based on feature similarity, we compute an associated geometric edge attribute that represents the minimum periodic distance between these atoms. This step ensures that the graph representation is invariant to equivalent unit-cell configurations, an essential requirement given that crystalline structures can have multiple equivalent representations due to symmetry.

Specifically, the calculation involves converting atomic coordinates into fractional coordinates to account explicitly for the periodic boundary conditions imposed by the lattice vectors:
\begin{equation}
\mathbf{d}_{\text{frac}} = \mathbf{L}^{-1}(\mathbf{r}_i - \mathbf{r}_j),
\end{equation}
where $\mathbf{d}_{\text{frac}}$ denotes the fractional distance vector between atoms $i$ and $j$. To identify the minimal periodic distance, we shift this fractional distance into the fundamental periodic cell by applying:
\begin{equation}
\mathbf{d}_{\text{frac}}^{\text{PBC}} = \mathbf{d}_{\text{frac}} - \lfloor \mathbf{d}_{\text{frac}} + 0.5 \rfloor,
\end{equation}
where $\lfloor \cdot \rfloor$ denotes the floor operation, effectively mapping the fractional coordinates into the range $[-0.5, 0.5]$. Finally, we convert the fractional minimal vector back into Cartesian coordinates:
\begin{equation}
\mathbf{d}_{\text{min}} = \mathbf{L}\mathbf{d}_{\text{frac}}^{\text{PBC}}.
\end{equation}

\begin{figure}[h]
    \centering
    \includegraphics[width=0.9\linewidth]{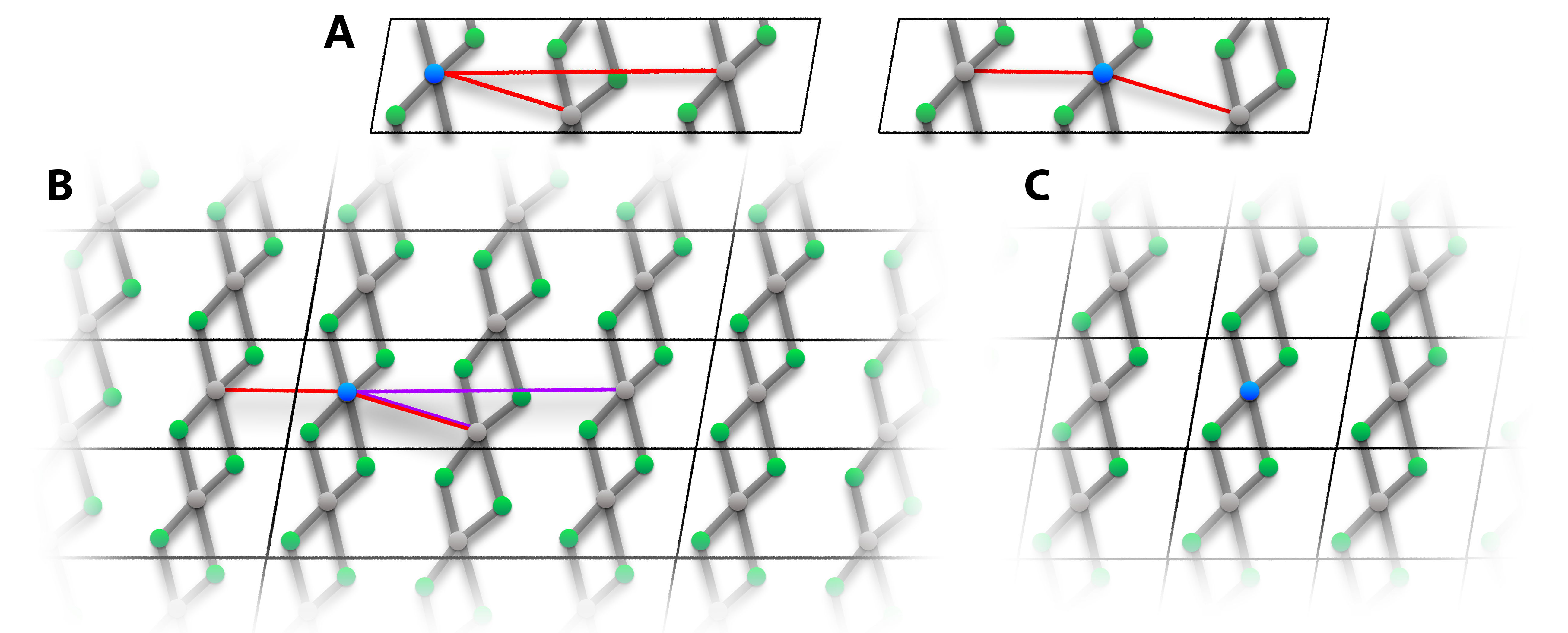}
    \caption{
    Illustration of the feature-space neighbourhood $\mathcal{G}_{\mathrm{feat}}$ under periodic boundary conditions. The neighbourhood is built around the blue reference atom and edges (red) link feature-similar atoms within the shown cell. \textbf{A} Two equivalent unit-cell choices lead to different similarity edges if periodicity is ignored, producing inconsistent graphs across equivalent cells. \textbf{B} Proposed periodic-invariant construction. Candidate neighbours (magenta) are first identified in feature space; for each candidate, we then select the minimal periodic image and add the corresponding edge (red), yielding consistent and invariant graphs. \textbf{C} Corner case in which no feature-similar atoms lie inside the reference unit cell. Similar atoms exist only as periodic replicas, resulting in an unconnected $\mathcal{G}_{\mathrm{feat}}$ graph and a vertically connected $\mathcal{G}_{\mathrm{atomistic}}$, which leads to poor information flow for this structure. Nevertheless, the Cell-Space Expert is able to propagate the information in such structures since the $\mathcal{G}_{\mathrm{cell}}$ graph is connected to the image replicas.
    }
    \label{fig:pbc_feat}
\end{figure}

This approach guarantees that the resulting geometric edge attributes consistently represent minimal distances under periodic boundary conditions and are invariant to different but equivalent unit-cell representations.  The detailed construction of $\mathcal{G}_{\mathrm{feat}}$ is visually illustrated in Figure~\ref{fig:pbc_feat}.

Due to the dynamic nature of the feature-space graph, where different edges may emerge or disappear between message-passing layers, edge features cannot be precomputed or reused. Thus, at each layer, we independently reconstruct the edges and re-encode their geometric attributes through a dedicated encoder named \textit{Feature-Edge Encoder}. This contrasts with the static edge encoders employed in other experts, ensuring accurate and consistent edge feature representations across the evolving graph topology.

The main limitation of this expert is that $\mathcal{G}_{\mathrm{feat}}$ is built with atoms restricted to a single unit cell. Some crystals exhibit no feature-level similarity within the cell but do exhibit similarity across periodic replicas. In such cases, $\mathcal{G}_{\mathrm{feat}}$ can become a disconnected graph and fail to improve message-passing propagation, as can be seen in Figure~\ref{fig:pbc_feat} \textbf{C}.

\subsection{Cell-Space Expert}

To address the limitations introduced by the Atomistic and Similarity experts in the message passing, such as the examples in Figure 2 B and Figure 3 C, we propose the Cell-Space expert. It captures long-range cell-to-cell interactions, such as delocalised electronic coupling, collective lattice correlations, and boundary-driven surface effects, which atom-level graphs cannot represent. The design encodes global periodic repetitions by introducing a single superatom node $s$. We initialise the embedding of $h_s$ with an attention-weighted sum of the atomic embeddings within the unit cell. See the Detailed Architecture section in the Supplementary Information for implementation details.

\begin{figure}[h]
    \centering
    \includegraphics[width=0.8\textwidth]{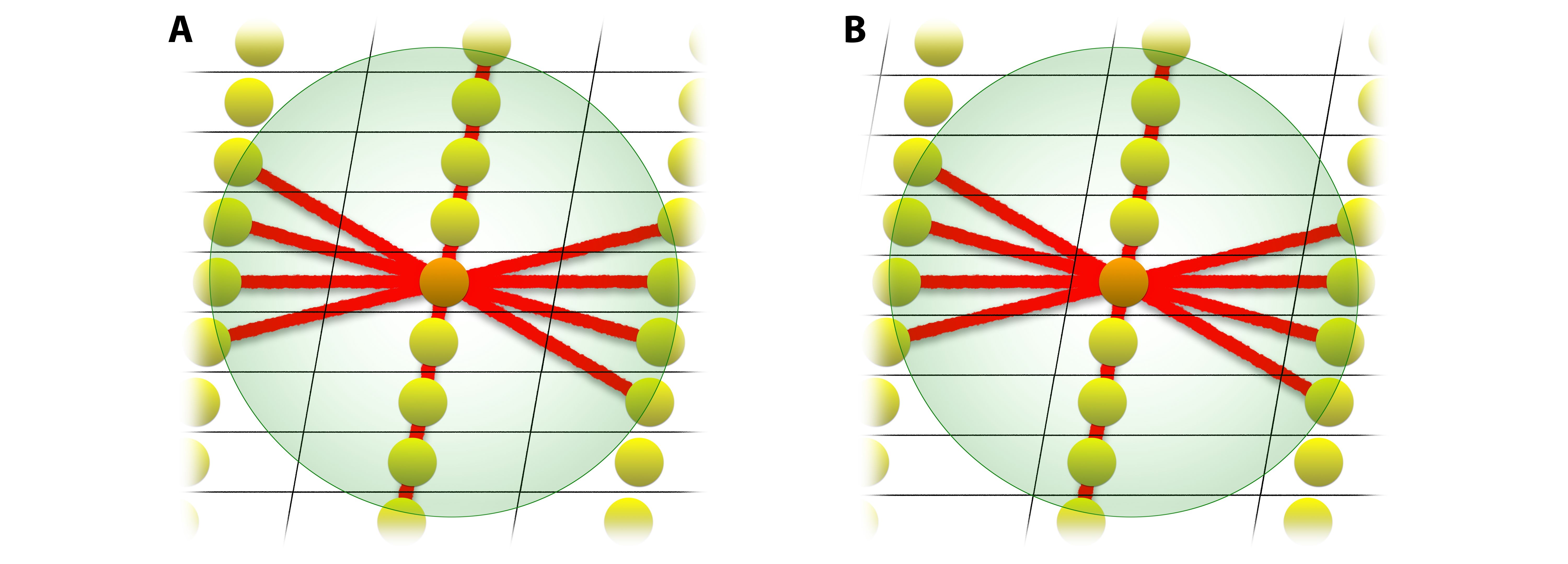}
    \caption{
    Cell-Space expert graph construction. The superatom (central orange node) connects to periodic replicas within a large cutoff $R_c$ (green circle), encoding lattice-scale periodicity through edges (red). \textbf{A–B} Two equivalent unit-cell choices produce the same set of replica connections because cell transformations preserve the relative distances and directions between the superatom and its periodic images. The only change comes from global rotations, which can be handled by defining an invariant cell~\cite{conformer}, using equivariant message-passing updates~\cite{conformer}, or applying rotational data augmentation~\cite{CARTNET}.
    }
    \label{fig:cell_graph}
\end{figure}

We construct a radius-based graph $\mathcal{G}_{\mathrm{cell}}$ around the superatom node with a cutoff radius $R_c$, significantly larger than the atomistic radius $r_c$ ($R_c \gg r_c$). Edges in this graph are initialised based on the geometric distances within this extended radius $R_c$ and are processed using an independent edge encoder. This edge encoder is analogous to the one used in the atomistic expert but tailored specifically to handle the larger cutoff radius. The significantly increased radius ensures that the graph explicitly captures lattice repetitions at the superatom scale. Figure \ref{fig:cell_graph} illustrates the detailed construction of the Cell-Space Expert graph $\mathcal{G}_{\mathrm{cell}}$. As discussed in Matformer~\cite{matformer} and iComformer~\cite{conformer}, the distance between the repetitions of the same atom is invariant to equivalent unit-cell transformations, and the rotational symmetries can be overcome thanks to defining an invariant cell~\cite{conformer}, equivariant message passing updates~\cite{conformer} or via rotational data augmentation~\cite{CARTNET}.

The Cell‐Space expert arises from the way electronic coupling between neighbouring cells varies with cell size. For small unit cells, atoms in adjacent images lie within the immediate chemical environment and therefore produce dense connectivity around the superatom. In contrast, for larger unit cells, significant cross‐cell interactions occur only for atoms near the boundaries, and the Atomistic Expert and the Similarity Expert already capture these boundary effects. In other words, for small unit cells, this expert is highly influential and forms many connections, whereas for larger cells, its reduced connectivity is compensated by more connections formed by the Atomistic and Similarity experts.

Other methods, such as Matformer~\cite{matformer}, also aim to overcome the limitations of relying only on the Atomistic experts. Matformer adds self-loops to each atom to encode distance features derived from the unique distances to each of the image repetitions. The main limitation is that, even though each atom is connected to its periodic image, the method does not capture intra-cell symmetries as effectively as the Similarity expert. In addition, every atom stores the same self-loop distances information, which introduces redundancy that scales with the number of atoms. Collapsing this shared information into a single superatom removes the redundancy and makes the approach more efficient while preserving the relevant periodic information. Our Cell-Space expert also scales negatively with the number of atoms in the unit cell. As the number of atoms increases, the typical cell size grows, periodic images move farther apart for a fixed cutoff $R_c$, and the superatom graph $\mathcal{G}_{\mathrm{cell}}$ contains fewer connections. Further comparisons on scalability and computational cost are provided in the Supplementary Information. The remaining design question for this expert is how to couple the superatom to the underlying atomic representations so that equivalent unit-cell transformations produce identical outputs.

\subsection{Multiscale Expert}

\begin{figure}[h]
    \centering
    \includegraphics[width=0.8\linewidth]{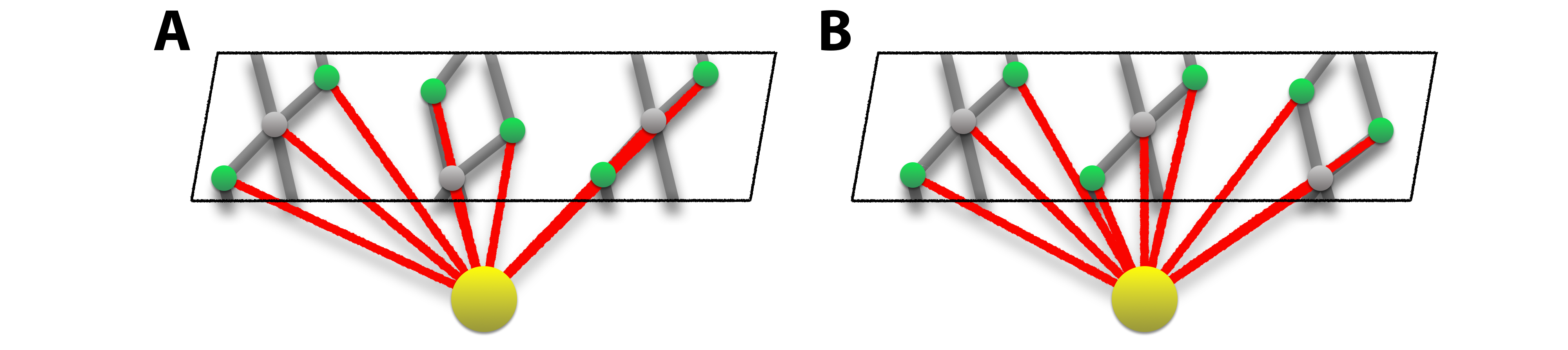}
    \caption{ Multiscale expert graph $\mathcal{G}_{\mathrm{multiscale}}$. The superatom node (yellow) forms a bipartite graph to all atoms in the unit cell, with edges shown in red. In this expert, message passing is feature-only: no distances, directions, or geometric information are used. \textbf{A} and \textbf{B} show two equivalent unit-cell representations of the same material. Because the connectivity depends only on the set of atomic embeddings and not on their coordinates, both representations yield the same graph and updates, making this expert invariant to unit-cell transformations.
    }
    \label{fig:multiscale}
\end{figure}

Finally, to capture interactions across distinct structural scales, we introduce the Multiscale expert, which explicitly connects the global superatom representation with atomic-level embeddings without breaking the invariance between equivalent unit-cell representations. The primary objective of this expert is to enable an efficient bidirectional information flow, allowing local atomic features to aggregate into a global representation and simultaneously distributing global contextual information back to the atomic nodes. Such coupling across scales is inspired by multiscale modeling frameworks commonly employed in computational physics~\cite{multiscale1,multiscale22}, where macroscale properties emerge from microscale interactions, and global constraints influence local states. Formally, we define the bipartite multiscale graph $\mathcal{G}_{\mathrm{multiscale}}$ as follows:

\begin{equation}
\mathcal{G}_{\mathrm{multiscale}} = \{ (s, i) \mid s \leftrightarrow i,\,\forall\, i \in \mathcal{V} \},
\end{equation}

where the superatom node $ s $ is bidirectionally connected to every atom $ i $ within the set of atomic nodes $\mathcal{V}$ representing the unit cell. The connectivity ensures that all atomic nodes share a direct path to exchange information with the superatom node, enforcing global coherence across scales. Figure \ref{fig:multiscale} visually illustrates the construction and connectivity of the multiscale graph $\mathcal{G}_{\mathrm{multiscale}}$. The message-passing operation between scales can be mathematically represented as follows:

\begin{equation}
\mathbf{h}_s^{(l+1)} = \mathbf{h}_s^{(l)} + \phi_{\mathrm{multiscale}}\left(\{\mathbf{h}_i^{(l)}\}_{i\in\mathcal{V}}\right).
\end{equation}

\begin{equation}
\mathbf{h}_i^{(l+1)} = \mathbf{h}_i^{(l)} + \phi_{\mathrm{multiscale}}\left(\mathbf{h}_s^{(l+1)}\right).
\end{equation}

Employing the same graph neural network transformation $\phi_{\mathrm{multiscale}}$ in both aggregation and distribution steps ensures consistency in multiscale information exchange.
The Multiscale expert is deliberately geometry-agnostic and operates only on the $N$ atomic embeddings from the unit cell. We modify the cross-scale message passing to exclude the edge features, which carry the geometric information. $\phi_{\mathrm{multiscale}}$ consumes only feature embeddings, bipartite edges carry no distances, angles, or relative-position encodings, and updates depend solely on $\{\mathbf{h}_i\}$ and $\mathbf{h}_s$. As illustrated in Figure~\ref{fig:multiscale}, each equivalent unit-cell transformation contains the same $N$ atoms. Their coordinates may vary under alternative unit-cell choices, yet the set of $N$ atomic embeddings remains the same.

Because the superatom has no defined spatial position and messages do not use geometric features, the construction preserves invariance under equivalent unit-cell transformations. Further details on these modifications are provided in the Detailed Architecture section of the Supplementary Information.

\subsection{Fusion of Experts}

To integrate complementary information captured by the different expert modules, we introduce a learned fusion mechanism at each aggregation step. The fusion strategy separately handles the superatom and atomic node embeddings, reflecting their distinct roles within the model.

For the superatom representation, we employ a gating mechanism controlled by a trainable parameter $\alpha$. This parameter, modulated through the sigmoid function $\sigma(\cdot)$, dynamically balances the contributions from the Cell-Space and Multiscale experts:

\begin{equation}
\mathbf{h}_s^{(l+1)} = \sigma(\alpha) \mathbf{h}_s^{(\mathrm{cell})} + \left(1 - \sigma(\alpha)\right) \mathbf{h}_s^{(\mathrm{multiscale})},
\end{equation}

where $\mathbf{h}_s^{(\mathrm{cell})}$ is the output superatom embedding from the Cell expert, $\mathbf{h}_s^{(\mathrm{multiscale})}$ is the output superatom embedding from the Multiscale expert, and $\sigma$ denotes the sigmoid activation function, allowing the model to adaptively favor either the global periodic information from the Cell-Space expert or the hierarchical interactions from the Multiscale expert, depending on the context.

For atomic node representations, we employ a separate fusion approach. Here, we define learnable parameters $\beta'$, $\gamma'$, and $\delta'$, corresponding to each expert module (Atomistic, Similarity, and Multiscale). These parameters are normalised via a softmax operation, producing weighting factors that sum to one, ensuring a meaningful convex combination of expert outputs:

\begin{equation}
[\beta, \gamma, \delta] = \text{Softmax}([\beta', \gamma', \delta']),
\end{equation}

\begin{equation}
\mathbf{h}_i^{(l+1)} = \beta \mathbf{h}_i^{(\mathrm{atomistic})} + \gamma \mathbf{h}_i^{(\mathrm{feat})} + \delta \mathbf{h}_i^{(\mathrm{multiscale})}.
\end{equation}

where $\mathbf{h}_i^{(\mathrm{atomistic})}$ is the output of the i-th atom embedding from the Atomistic expert, $\mathbf{h}_i^{(\mathrm{feat})}$ is the output of the i-th atom embedding from the Similarity expert, and $\mathbf{h}_i^{(\mathrm{multiscale})}$ is the output of the i-th atom embedding from the Multiscale expert.

This weighted fusion allows each atomic embedding to optimally integrate geometric, chemical, and scale-dependent information provided by the distinct expert modules, ensuring the representation remains flexible and expressive across diverse crystal structures.

\section{Results}
\label{sec:results}

To rigorously evaluate the effectiveness of our proposed PRISM model, we conduct extensive experiments using three widely-adopted crystal property benchmarks: JARVIS~\cite{Jarvis}, the Materials Project~\cite{megnet}, and MatBench~\cite{dunn2020benchmarking}. These benchmarks cover a broad range of crystalline materials and property prediction tasks, varying significantly in dataset scale from large-scale tasks comprising up to 69,239 crystals, medium-scale tasks with approximately 18,171 crystals, to small-scale tasks consisting of only 5,450 crystals. In particular, the MatBench dataset includes diverse tasks such as \textit{e\_form}, comprising 132,752 crystals, and \textit{jdft2d}, consisting of 636 two-dimensional (2D) crystal structures. Further details from the datasets can be found in the Dataset Description Section from the Supplementary Information. 

We follow standardised experimental settings consistent with prior works, proposed by Keqiang et al.~\cite{matformer}, using Mean Absolute Error (MAE) as our evaluation metric to facilitate direct performance comparisons, but also adding the standard deviation between different initialisation seeds to show the robustness of the method, as proposed by \cite{CARTNET}. Additionally, we compare PRISM with several state-of-the-art baseline methods, including CGCNN~\cite{CGCNN}, Schnet~\cite{schnet}, MEGNET~\cite{megnet}, GATGNN~\cite{CATGNN}, ALIGNN~\cite{alignn}, MatFormer~\cite{matformer}, PotNet~\cite{potnet}, M3GNet~\cite{chen2022graph}, MODNet~\cite{dunn2020benchmarking}, coGN~\cite{cogn}, eComFormer~\cite{conformer}, iComFormer~\cite{conformer}, and CartNet~\cite{CARTNET}. 
Based on the results of the Table~\ref{tab:ablations_agnostic} from the Discussion Section, in all our experiments we used CartNet~\cite{CARTNET} as expert module. The Detailed Architecture Section from the Supplementary Information provides detailed implementation information for all the architecture used.
Additional details about the computational details, PRISM's hyperparameters, and training configurations can be found in the Training Details Section from the Supplementary Information.

\paragraph{Jarvis Dataset}

Table~\ref{tab:mae-results} shows that among the tested architectures, PRISM delivers the highest performance in every task. In detail, our method improves formation energy error by about 4.36\%, enhances band gap (OPT) accuracy by roughly 5.24\%, and lowers total energy error by nearly 0.9\% compared with the runner up. It also reduces band gap (MBJ) error by approximately 6.54\% and cuts Ehull error by close to 45.2\%.

\begin{table}[h]
\centering
\caption{MAE results for the different tested architectures in the test split from the JARVIS dataset. 
         The best result is in \textbf{bold} and the second-best is \underline{underlined}. 
         }
\label{tab:mae-results}
\resizebox{\linewidth}{!}{%
\begin{tabular}{lccccc}
\toprule
 & 
\textbf{Form. Energy}  & 
\textbf{Band gap (OPT)} & 
\textbf{Total energy} & 
\textbf{Band gap (MBJ)}  & 
\textbf{Ehull}  \\
\textbf{Method} & 
 (meV/atom)  & 
 (meV)   & 
 (meV/atom)   & 
 (meV)   & 
 (meV/atom)  \\
\midrule
CGCNN      & 63  & 200 & 78 & 410 & 170 \\
SchNet     & 45  & 190 & 47 & 430 & 140 \\
MEGNET     & 47  & 145 & 58 & 340 & 84  \\
GATGNN     & 47  & 170 & 56 & 510 & 120 \\
ALIGNN     & 33  & 142 & 37 & 310 & 76  \\
Matformer  & 32.5           & 137           & 35             & 300           & 64             \\
PotNet    & 29.4           & 127           & 32             & 270          & 55             \\
eComFormer & 28.4         &  124           & 32             &  280    &  {\ul 44}       \\
iComFormer &   27.2     & {\ul 122}    & {\ul 28.8}     & {\ul260} &  47            \\
CartNet      & {\ul 27.05 $\pm$ 0.07}  &  {\ul 115.31 $\pm$ 3.36 } & {\ul 26.58 $\pm$ 0.28} & {\ul 253.03 $\pm$ 5.20 }       &  {\ul 43.90 $\pm$ 0.36} \\
PRISM           & \textbf{25.87 $\pm$ 0.36} & \textbf{109.26 $\pm$ 2.54} & \textbf{26.34 $\pm$ 0.38} & \textbf{236.49 $\pm$  5.56} & \textbf{24.07 $\pm$ 0.62}  \\
\bottomrule
\end{tabular}
}
\end{table}

\paragraph{Materials Project Dataset}

As shown in Table~\ref{tab:results-megnet}, among highly competitive baselines PRISM achieves the best performance for formation energy, band gap and bulk modulus. Specifically, our approach reduces the formation energy error by about 5.0\% and the band gap error by nearly 5.8\% compared with the second best method. In bulk modulus prediction PRISM attains the top result, while in shear modulus it shows an error that is around 2.8\% higher than the best value.

\begin{table}[h]
\centering
\caption{MAE results for the different tested architectures in the test split from the Material Project Dataset. Best result in \textbf{bold} and second-best {\ul underlined}. }
\label{tab:results-megnet}
\resizebox{\linewidth}{!}{%
\begin{tabular}{lcccc}
\toprule
 &
  \textbf{Form. Energy}  &
  \textbf{Band Gap}  &
  \textbf{Bulk Moduli}  &
  \textbf{Shear Moduli}  \\
\textbf{Method} &
  (meV/atom) &
  (meV) &
  (log(GPa)) &
  (log(GPa)) \\
\midrule
CGCNN      & 31  & 292 & 0.047  & 0.077 \\
SchNet     & 33  & 345 & 0.066  & 0.099 \\
MEGNET     & 30  & 307 & 0.060  & 0.099 \\
GATGNN     & 33  & 280 & 0.045  & 0.075 \\
ALIGNN     & 22  & 218 & 0.051  & 0.078 \\
Matformer  & 21             & 211          & 0.043  & 0.073  \\
PotNet    & 18,8           & 204          & 0.04   & {\ul 0.065}  \\
eComFormer & 18.16 & 202          & 0.0417 & 0.0729 \\
iComFormer &  18.26    & 193 & {\ul 0.038} & \textbf{0.0637} \\
CartNet      & {\ul 17.47 $\pm$ 0.38}       &  {\ul 190.79 $\pm$ 3.14}    & \textbf{0.033 $\pm$ 0.94 $\mathbf{\cdot 10^{-3}}$}     &  $\mathbf{0.0637 \pm 0.0008}$ \\
PRISM &  \textbf{ 16.59 $\pm$ 0.10} & \textbf{179.71 $\pm$ 1.58} & \textbf{ 0.033 $\pm$ 1.17} $\mathbf{\cdot 10^{-3}}$ & {\ul 0.0655 $\pm$ 0.0008} \\
\bottomrule
\end{tabular}%
}
\end{table}

\paragraph{MatBench Dataset}

Table~\ref{tab:matbench_comparison} compares PRISM against state-of-the-art baselines on two MatBench tasks. On the e\_form benchmark (132\,752 samples), PRISM achieves the lowest MAE and RMSE (15.20\,$\pm$\,0.31, 30.43\,$\pm$\,1.38 meV/atom), improving upon the next-best models (eComFormer and iComFormer at 16.5\,$\pm$\,0.3 meV/atom). For jdft2d (636 samples), PRISM attains an MAE of 38.41\,$\pm$\,12.44 GPa and RMSE of 97.90\,$\pm$\,38.25 GPa, closely following iComFormer’s top MAE (34.8\,$\pm$\,9.9 GPa) and RMSE (96.1\,$\pm$\,46.3 GPa) while outperforming other methods. We note that jdft2d comprises only 636 two-dimensional crystals; this low-data regime is likely not enough to achieve the mixture of experts full potential but still achieve a competitive result. The significant error deviation in the metrics suggests that additional labelled data or targeted augmentation could yield further gains. These results demonstrate PRISM’s superior accuracy in large-scale formation-energy predictions and its competitive performance under scarce-data scenarios.

\begin{table}[h]
\centering

\caption{MAE and RMSE with mean and std from all the test splits from the Matbench dataset. Best result in \textbf{bold} and second-best {\ul underlined}.}
\label{tab:matbench_comparison}
\begin{tabular}{lcccccc}
\toprule
 & \multicolumn{2}{c}{\textbf{e\_form} (meV)} & & \multicolumn{2}{c}{\textbf{jdft2d} (GPa)}  \\
\cmidrule{2-3} \cmidrule{5-6}
\textbf{Method} & MAE  & RMSE  && MAE  & RMSE  \\
\midrule
MODNet       & 44.8 $\pm$ 3.9  & 88.8 $\pm$ 7.5  && \textbf{33.2 $\pm$ 7.3}  & \underline{96.7 $\pm$ 40.4} \\
ALIGNN     & 21.5 $\pm$ 0.5  & 55.4 $\pm$ 5.5  && 43.4 $\pm$ 8.9          & 117.4 $\pm$ 42.9 \\
coGN       & 17.0 $\pm$ 0.3  & 48.3 $\pm$ 5.9  && 37.2 $\pm$ 13.7         & 101.2 $\pm$ 55.0 \\
M3GNet     & 19.5 $\pm$ 0.2  & -              && 50.1 $\pm$ 11.9         & - \\
eComFormer  & \underline{16.5 $\pm$ 0.3}  & 45.4 $\pm$ 4.7 && 37.8 $\pm$ 9.0          & 102.2 $\pm$ 46.4 \\
iComFormer  & \underline{16.5 $\pm$ 0.3}  & \underline{43.8 $\pm$ 3.7} && \underline{34.8 $\pm$ 9.9} & \textbf{ 96.1 $\pm$ 46.3} \\
PRISM & \textbf{15.20 $\pm$ 0.31} & \textbf{30.43 $\pm$ 1.38} & & 38.41 $\pm$ 12.44  & 97.90 $\pm$ 38.25 & \\
\bottomrule
\end{tabular}

\end{table}

\section{Discussion}
\label{sec:discussion}

\subsection{Effect of the expert modules}
We begin by quantifying how each specialist in PRISM contributes to accuracy and efficiency. The aim is to understand which cues, namely local atomistic geometry, feature-space similarity that respects periodicity, cross-scale aggregation, and explicit cell-scale periodicity, drive improvements on each target, and whether combining them yields complementary gains.

Table~\ref{tab:ablations} presents a controlled progression from an atomistic-only baseline to the full mixture. Starting from CartNet with only an atomistic radius graph (formation energy $27.05\pm0.07$\,meV/atom and Ehull $43.90\pm0.36$\,meV/atom), adding the Similarity expert already lowers Ehull to $34.91\pm1.28$\,meV/atom with a moderate increase in time per epoch. Enforcing periodic invariance in feature space further improves both targets, reaching $26.89\pm0.19$\,meV/atom for formation energy and $32.42\pm0.51$\,meV/atom for Ehull at the same parameter count. The Multiscale expert on its own reduces formation-energy error to $26.92\pm0.17$\,meV/atom, consistent with its role in aggregating atomic information without introducing new geometric edges. Adding the Cell-Space expert on top of Multiscale strengthens formation energy to $26.05\pm0.21$\,meV/atom, showing the benefit of explicit lattice-scale periodic cues. The best overall performance appears when all experts are fused: $25.87\pm0.36$\,meV/atom for formation energy and $24.07\pm0.62$\,meV/atom for Ehull. Together, these results indicate that feature-space similarity under periodic boundary conditions is especially effective for Ehull, while Multiscale and Cell-Space contribute strongly to formation energy; their combination is needed to reach the highest accuracy.

\begin{table}[h]
  \caption{\textbf{Effect of expert modules on JARVIS.} Checkmarks indicate which specialists are enabled (Atomistic, Similarity/Feat., Multiscale, and Cell). $\checkmark_{\mathrm{PBC}}$ denotes feature-space similarity made periodic-invariant via minimum-distance edges. We report mean~$\pm$~s.d. over seeds for formation energy and Ehull (meV/atom), time per epoch (s), and parameter count (M). Best and second-best per column are highlighted in \textbf{bold} and \underline{underline}, respectively.}
  \label{tab:ablations}
  \centering
  \resizebox{\linewidth}{!}{
  \begin{tabular}{cccccccc}
      \toprule
      \textbf{Atom.} & \textbf{Feat.} & \textbf{Multi.} & \textbf{Cell} & \textbf{Form.\ Energy} & \textbf{Ehull} & \textbf{Time/epoch} & \textbf{\#Params} \\
      & & & & (meV/atom) & (meV/atom) & (s) & \\
      \midrule
      \checkmark &  &  &  & 27.05 $\pm$ 0.07 & 43.90 $\pm$ 0.36 & \textbf{8 $\pm$ 0.05} & \textbf{2.5M} \\
      \checkmark & \checkmark &  &  & 27.10 $\pm$ 0.27 & 34.91 $\pm$ 1.28 & \underline{16.46 $\pm$ 0.13} & 4.9M \\
      \checkmark & $\checkmark_{PBC}$ &  &  & 26.89 $\pm$ 0.19 & \underline{32.42 $\pm$ 0.51} & 16.83 $\pm$ 0.12 & 4.9M \\
      \checkmark &  & \checkmark &  & 26.92 $\pm$ 0.17 & 39.96 $\pm$ 0.37 & 19.28 $\pm$ 0.20 & \underline{4.5M} \\
      \checkmark &  & \checkmark & \checkmark & \underline{26.05 $\pm$ 0.21} & 38.88 $\pm$ 1.10 & 26.61 $\pm$ 0.20 & 6.6M \\
      \checkmark & $\checkmark_{PBC}$ & \checkmark & \checkmark & \textbf{25.87 $\pm$ 0.36} & \textbf{24.07 $\pm$ 0.62} & 37.71 $\pm$ 0.29 & 9M \\
      \bottomrule
  \end{tabular}
  }
\end{table}

In summary, the mixture consistently improves both targets relative to the atomistic-only baseline, with a controlled growth in parameters and time per epoch. This supports the decision to integrate short-range geometry, periodic-invariant similarity, cell-level periodicity, and cross-scale aggregation within one model.

\subsection{Backbone-agnostic behaviour and choice of message passing}
We next test whether these gains are consistent across message-passing backbones and use this comparison to choose the backbone for the main experiments. Beyond consistency, we also test whether the improvements can be explained simply by increasing parameter count, depth, width, or radius, or whether the mixture itself contributes additional signal.

Table~\ref{tab:ablations_agnostic} compares vanilla eComformer, iComformer, and CartNet variants against the same backbones augmented with all PRISM experts. For eComformer, the formation-energy error decreases from $28.40$ to $27.74$\,meV/atom and Ehull decreases from $44.00$ to $31.73$\,meV/atom when the experts are added. For iComformer, formation energy decreases from $27.2$ to $26.46$\,meV/atom and Ehull decreases from $47$ to $36.06$\,meV/atom. With CartNet as the backbone, PRISM attains the strongest results overall, reaching $25.87\pm0.36$\,meV/atom for formation energy and $24.07\pm0.62$\,meV/atom for Ehull.

\begin{table}[h]
\caption{Ablation study of different state-of-the-art architectures used as experts on formation energy and Ehull on the JARVIS dataset. Best is in \textbf{bold} and second-best is \underline{underlined}.}
\label{tab:ablations_agnostic}
\centering
\resizebox{\linewidth}{!}{
\begin{tabular}{lcccc}
\toprule
 & \textbf{Form. energy}  & \textbf{Ehull} & \textbf{Time/epoch} & \textbf{\#Params} \\
\textbf{Architecture} &  (meV/atom) & (meV/atom) & (seconds) & \\
\midrule
eComformer (vanilla)        & $28.40$                 & $44.00$                 & $\underline{23.87 \pm 0.88}$ & $12.4\,\text{M}$ \\
eComformer (512 dim)        & $31.04 \pm 0.49$        & $44.99 \pm 2.74$        & $40.11 \pm 0.66$            & $16.8\,\text{M}$ \\
eComformer (16 layers)      & $29.53 \pm 1.78$        & $42.35 \pm 5.03$        & $73.63 \pm 2.20$            & $16.6\,\text{M}$ \\
+ all PRISM experts         & $27.74 \pm 0.17$        & $\underline{31.73 \pm 2.69}$ & $42.65 \pm 1.07$        & $15\,\text{M}$ \\
\midrule
iComformer (vanilla)        & $27.2$                  & $47$                     & $29.90 \pm 0.38$            & $\underline{5\,\text{M}}$ \\
iComformer (512 dim)        & $29.25 \pm 0.61$        & $43.73 \pm 1.55$         & $62.18 \pm 0.17$            & $25\,\text{M}$ \\
iComformer (16 layers)      & $31.96 \pm 2.08$        & $75.48 \pm 8.39$         & $72.80 \pm 1.31$            & $16.5\,\text{M}$ \\
+ all PRISM experts         & $\underline{26.46 \pm 0.62}$ & $36.06 \pm 0.79$     & $41.29 \pm 1.32$            & $15\,\text{M}$ \\
\midrule
CartNet (vanilla)           & $27.05 \pm 0.07$        & $43.90 \pm 0.36$         & $\mathbf{8} \pm \mathbf{0.05}$ & $\mathbf{2.5}\,\text{\textbf{M}}$ \\
CartNet (10 \AA)            & $27.61 \pm 2.47$        & $32.86 \pm 1.58$         & $82.58 \pm 0.16$            & $\mathbf{2.5}\,\text{\textbf{M}}$ \\
CartNet (16 layers)         & $28.91 \pm 0.88$        & $34.44 \pm 3.39$         & $49.64 \pm 0.95$            & $8.8\,\text{M}$ \\
CartNet (512 dim)           & $27.18 \pm 0.22$        & $41.39 \pm 1.31$         & $27.38 \pm 0.19$            & $9.8\,\text{M}$ \\
+ all PRISM experts                       & $\mathbf{25.87} \pm \mathbf{0.36}$ & $\mathbf{24.07} \pm \mathbf{0.62}$ & $37.71 \pm 0.29$ & $9\,\text{M}$ \\
\bottomrule
\end{tabular}
}
\end{table}

Importantly, the backbone comparisons show that the gains are not a by-product of simply making models larger or extending the cutoff. Wider or deeper variants (512-dimensional or 16 layers) carry more parameters but do not match the accuracy achieved by the mixture with experts. Likewise, CartNet with a larger cutoff of 10\,\AA\ has the same parameter count as the baseline, increases time per epoch substantially, and still underperforms PRISM. These controls indicate that the mixture of experts contributes complementary information that is not recovered by scaling depth, width, or radius alone.

From an efficiency perspective, PRISM with CartNet runs at $37.71\pm0.29$\,s/epoch with 9M parameters, which is competitive with other state-of-the-art baselines such as vanilla eComformer ($23.87\pm0.88$\,s/epoch) and iComformer ($29.90\pm0.38$\,s/epoch), while delivering a considerable improvement in accuracy. When the experts are added to these backbones, the training time increases modestly: iComformer rises by about $10$\,s/epoch to $41.29\pm1.32$\,s/epoch, and eComformer by about $20$\,s/epoch to $42.65\pm1.07$\,s/epoch. For CartNet the overhead is larger and scales nearly linearly with the number of experts. CartNet uses very simple input encoders and concentrates most of the computation in the message-passing updates, which makes the vanilla model extremely fast but also means that duplicating message-passing modules across experts increases cost almost proportionally. This is why PRISM with CartNet is on the order of four times slower than vanilla CartNet, yet it remains competitive with the other baselines even without the experts. Given the consistent gains across all backbones and the favorable accuracy–efficiency balance, we adopt CartNet as the message-passing core in the remainder of the study. The pattern observed across eComformer, iComformer, and CartNet shows that the proposed mixture is largely agnostic to the choice of message-passing architecture.

\subsection{Interpretability via fusion–weight visualisation}

\begin{figure}[h]
\centering
\includegraphics[width=\linewidth]{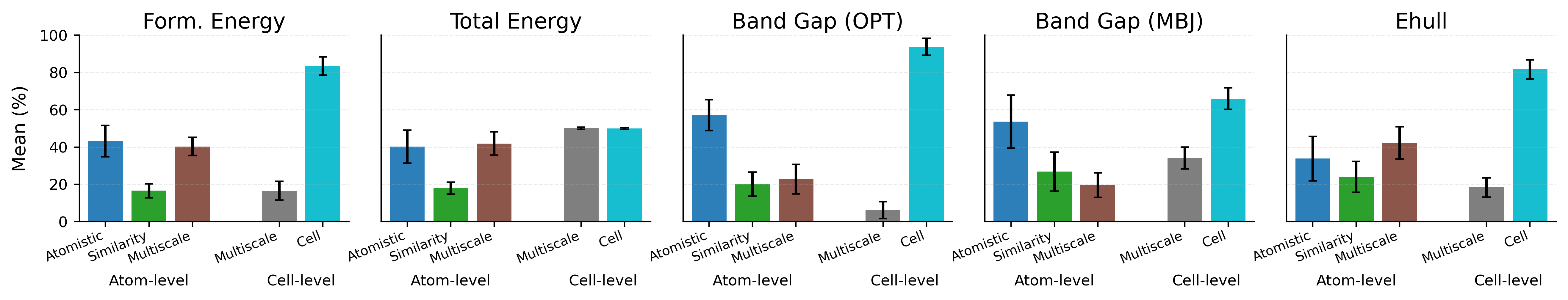}
\caption{\textbf{Fusion–weight analysis.} Average fusion weights over layers and seeds on the JARVIS test set. Each subpanel shows atom–level (\textit{Multiscale}, \textit{Atomistic}, \textit{Similarity}) and cell–level (\textit{Multiscale}, \textit{Cell}) contributions for a single target property.}
\label{fig:fusion-weights}
\end{figure}

We examine how PRISM distributes responsibility across experts and whether these distributions reflect the chemistry and physics of each target. To do so, we summarise the learned fusion weights by averaging across layers and random seeds, and then analyse the resulting allocations to the \textit{Atomistic}, \textit{Similarity}, \textit{Multiscale} and \textit{Cell} experts. Figure~\ref{fig:fusion-weights} reports the mean fusion weights for each property, separated into atom–level and cell–level contributions.

To ground the discussion, we briefly recall what each target represents. The Formation Energy is the energy per atom of a compound relative to the isolated neutral atoms; it is a crucial parameter for quantifying thermodynamic stability. Total Energy is the ground–state energy of the relaxed cell. Band Gap (OPT) is the Kohn–Sham gap computed with the OPT (optB88-vdW) functional~\cite{opt88}. Band Gap (MBJ) applies a modified exchange functional that includes an additional term proposed by Becke–Johnson~\cite{MBJ}, which mitigates the underestimation of gaps by Generalised-Gradient Approximation (GGA)~\cite{gga} functionals. Finally, Ehull (energy above the convex hull) measures the distance of a material from the stability hull defined by all competing phases of the same composition; stable entries have Ehull=0.

Two robust patterns are evident in Figure~\ref{fig:fusion-weights}. First, for properties with a strong global or band–structure character the cell–level pathway dominates: Band Gap (OPT) and Ehull  place most weight on the \textit{Cell} expert, and Band Gap (MBJ) still prefers \textit{Cell} over cell–level \textit{Multiscale}. Second, within atom–level fusion, energy–like targets favour \textit{Atomistic} and \textit{Multiscale} approaches, whereas the \textit{Similarity} expert becomes more prominent for electronic and stability descriptors.

Although both targets are reported per atom in JARVIS, the model allocates weights differently to formation energy and total energy. For formation energy, the atom-level fusion is dominated by \textit{Atomistic} and \textit{Multiscale} share (see Figure~\ref{fig:fusion-weights}), because formation energy is built as a difference between the compound and its elemental references; this reference normalisation cancels much of the atomic contributions, leaving local bonding, coordination, hybridisation and short interatomic distances as the primary determinants. By contrast, total energy per atom is an absolute quantity that still aggregates electrostatic and dispersion effects into the energy density experienced by each site; dividing by atom count removes trivial size scaling but not the underlying long-range couplings. Consequently, PRISM assigns a larger share to the cell-level \textit{Multiscale} pathway for total energy, where the superatom aggregator pools information across multiple neighbourhood radii to capture these extended interactions.

The behaviour of the band gap in the Figure~\ref{fig:fusion-weights} further illustrates the role of scale. With OPT, the gap is primarily governed by global band–structure features, such as periodicity, crystal symmetry, and lattice parameters; therefore, the model assigns overwhelming weight to the \textit{Cell} expert. Introducing MBJ changes the picture. MBJ raises conduction bands by including an additional kinetic–energy–density–dependent potential in the exchange term of the functional. This correction improves both periodic gap properties and the local chemical description (orbital energies, p–d hybridisation, and crystal–field splittings). This increases the relative contribution of atom–level paths, notably \textit{Similarity} and \textit{Multiscale}, compared with OPT, even though \textit{Cell} remains the single most significant contributor. Importantly, this shift is not primarily due to van der Waals dispersion, since OPT already includes a dispersion correction (optB88-vdW). The additional multiscale weight with MBJ reflects the need to propagate the proper local electronic structure to a long-range level and to capture subtle short–to–mid–range electronic effects beyond a pure cell–level summary.

The weights for Ehull shown in the Figure~\ref {fig:fusion-weights} are also physically reasonable. Because Ehull compares a structure’s formation energy with all competing phases of the same composition, it is strongly influenced by stoichiometry, density and lattice metrics. This explains the pronounced weight on the \textit{Cell} expert. At the same time, near–degenerate polymorphs and ordering motifs can shift a compound on or off the hull; capturing these subtle differences benefits from a \textit{Multiscale} expert, together with a smaller \textit{Atomistic}/\textit{Similarity} contribution.

Taken together, the property–aware allocation of weights accords with chemical and physical intuition. The fusion layer therefore improves interpretability in two complementary ways: it reveals which structural scale (atomic, similarity, cell or multiscale) is most informative for each property, and it enables the prediction to be decomposed into expert–wise contributions. As an additional quantitative check, correlations between mean weights and interpretable descriptors, such as local coordination for \textit{Atomistic}, dispersion and sublattice identities for \textit{Similarity}, lattice translation for \textit{Cell}, and long-range effects for \textit{Multiscale}, support a mechanistic interpretation rather than using a non-physically informed model.

\section{Conclusion, Limitations, and Future Work}

In this work, we introduce PRISM, a novel ensemble approach that leverages multiple graph neural network experts, designed to explicitly encode multiscale interactions and feature similarities with periodic boundary conditions for crystalline materials. Evaluated extensively on widely-used benchmarks, including JARVIS, Materials Project, and MatBench, our model achieved state-of-the-art performance across most tested properties, demonstrating its effectiveness in capturing the complex structural and chemical features of crystals.

A key limitation of this work is its exclusive focus on periodic crystal structures, which restricts its direct applicability to tasks that lack inherent periodicity, such as molecular property prediction. Moreover, the current similarity expert encodes only a single unit cell, leaving infinite lattice repetitions insufficiently represented, while the multiscale module considers only cell-level repetition frequencies rather than broader coarse-grained approaches. Finally, our approach focuses on unit cell representation; the usage of other cell representations, such as supercells, is compatible, but it would require retraining. Notwithstanding these limitations, PRISM offers clear strengths: it explicitly encodes unit-cell periodicity in both geometry and feature space, fuses local and global information through a multiscale mixture-of-experts design (Atomistic, Similarity, Multiscale, and Cell), and achieves state-of-the-art accuracy on JARVIS, Materials Project, and MatBench benchmarks, surpassing competitive baselines across most targets. Future work could adapt PRISM for molecular contexts by incorporating structural differences, integrating larger supercells or infinite lattice considerations into the feature graph, and introducing coarse-grained experts to enhance multiscale modelling for more complex hierarchical materials.

\section{Code availability}

The code to recreate the results of the paper can be found at: (the code will be released on GitHub upon acceptance).

\bibliography{refs}

@Article{CARTNET,
  author  = {Sol{\'e}, {\`A}lex and Mosella-Montoro, Albert and Cardona, Joan and G{\'o}mez-Coca, Silvia and Aravena, Daniel and Ruiz, Eliseo and Ruiz-Hidalgo, Javier},
  title   = {A Cartesian encoding graph neural network for crystal structure property prediction: application to thermal ellipsoid estimation},
  journal = {Digital Discovery},
  year    = {2025},
  volume  = {4},
  number  = {3},
  pages   = {694--710},
  publisher = {RSC},
  doi     = {10.1039/D4DD00352G},
  url     = {http://dx.doi.org/10.1039/D4DD00352G},
}

@inproceedings{conformer,
  title={Complete and Efficient Graph Transformers for Crystal Material Property Prediction},
  author={Yan, Keqiang and Fu, Cong and Qian, Xiaofeng and Qian, Xiaoning and Ji, Shuiwang},
  booktitle={International Conference on Learning Representations},
  year={2024}
}

@article{matformer,
  title={Periodic graph transformers for crystal material property prediction},
  author={Yan, Keqiang and Liu, Yi and Lin, Yuchao and Ji, Shuiwang},
  journal={Advances in Neural Information Processing Systems},
  volume={35},
  pages={15066--15080},
  year={2022}
}

@inproceedings{potnet,
  title={Efficient approximations of complete interatomic potentials for crystal property prediction},
  author={Lin, Yuchao and Yan, Keqiang and Luo, Youzhi and Liu, Yi and Qian, Xiaoning and Ji, Shuiwang},
  booktitle={International Conference on Machine Learning},
  pages={21260--21287},
  year={2023},
  organization={PMLR}
}

@article{dunn2020benchmarking,
  title={Benchmarking materials property prediction methods: the Matbench test set and Automatminer reference algorithm},
  author={Dunn, Alexander and Wang, Qi and Ganose, Alex and Dopp, Daniel and Jain, Anubhav},
  journal={npj Computational Materials},
  volume={6},
  number={1},
  pages={138},
  year={2020},
  publisher={Nature Publishing Group},
  doi={10.1038/s41524-020-00406-3}
}

@Article{cogn,
author ="Ruff, Robin and Reiser, Patrick and Stühmer, Jan and Friederich, Pascal",
title  ="Connectivity optimized nested line graph networks for crystal structures",
journal  ="Digital Discovery",
year  ="2024",
volume  ="3",
issue  ="3",
pages  ="594-601",
publisher  ="RSC",
doi  ="10.1039/D4DD00018H",
url  ="http://dx.doi.org/10.1039/D4DD00018H",}

@article{chen2022graph,
  title={Graph networks as a universal machine learning framework for molecules and crystals},
  author={Chen, Chi and Ye, Weike and Zuo, Yunxing and Zheng, Chen and Ong, Shyue Ping},
  journal={Nature Machine Intelligence},
  volume={4},
  number={9},
  pages={761--770},
  year={2022},
  publisher={Nature Publishing Group},
  doi={10.1038/s43588-022-00349-3}
}

@inproceedings{schnet,
 author = {Sch\"{u}tt, Kristof and Kindermans, Pieter-Jan and Sauceda Felix, Huziel Enoc and Chmiela, Stefan and Tkatchenko, Alexandre and M\"{u}ller, Klaus-Robert},
 booktitle = {Advances in Neural Information Processing Systems},
 editor = {I. Guyon and U. Von Luxburg and S. Bengio and H. Wallach and R. Fergus and S. Vishwanathan and R. Garnett},
 pages = {},
 publisher = {Curran Associates, Inc.},
 title = {SchNet: A continuous-filter convolutional neural network for modeling quantum interactions},
 url = {https://proceedings.neurips.cc/paper_files/paper/2017/file/303ed4c69846ab36c2904d3ba8573050-Paper.pdf},
 volume = {30},
 year = {2017}
}

@article{alignn,
  title={Benchmarking graph neural networks for materials chemistry},
  author={Tran, Kevin and Yao, Zhenwei and Batzner, Simon and Sun, Jiahua and Kornbluth, Mordechai and Kusne, Agha and Shapeev, Alexander and DiStasio Jr, Robert A and Marzari, Nicola and others},
  journal={npj Computational Materials},
  volume={8},
  number={1},
  pages={96},
  year={2022},
  publisher={Nature Publishing Group},
  doi={10.1038/s41524-021-00650-1}
}

@Article{CATGNN,
author ="Louis, Steph-Yves and Zhao, Yong and Nasiri, Alireza and Wang, Xiran and Song, Yuqi and Liu, Fei and Hu, Jianjun",
title  ="Graph convolutional neural networks with global attention for improved materials property prediction",
journal  ="Phys. Chem. Chem. Phys.",
year  ="2020",
volume  ="22",
issue  ="32",
pages  ="18141-18148",
publisher  ="The Royal Society of Chemistry",
doi  ="10.1039/D0CP01474E",
url  ="http://dx.doi.org/10.1039/D0CP01474E",}

@article{megnet,
author = {Chen, Chi and Ye, Weike and Zuo, Yunxing and Zheng, Chen and Ong, Shyue Ping},
title = {Graph Networks as a Universal Machine Learning Framework for Molecules and Crystals},
journal = {Chemistry of Materials},
volume = {31},
number = {9},
pages = {3564-3572},
year = {2019},
doi = {10.1021/acs.chemmater.9b01294},
URL = {  https://doi.org/10.1021/acs.chemmater.9b01294
},
eprint = { 
https://doi.org/10.1021/acs.chemmater.9b01294    
}
}

@article{CGCNN,
  title = {Crystal Graph Convolutional Neural Networks for an Accurate and Interpretable Prediction of Material Properties},
  author = {Xie, Tian and Grossman, Jeffrey C.},
  journal = {Phys. Rev. Lett.},
  volume = {120},
  issue = {14},
  pages = {145301},
  numpages = {6},
  year = {2018},
  month = {Apr},
  publisher = {American Physical Society},
  doi = {10.1103/PhysRevLett.120.145301},
  url = {https://link.aps.org/doi/10.1103/PhysRevLett.120.145301}
}

@InProceedings{SPOTR,
    author    = {Park, Jinyoung and Lee, Sanghyeok and Kim, Sihyeon and Xiong, Yunyang and Kim, Hyunwoo J.},
    title     = {Self-Positioning Point-Based Transformer for Point Cloud Understanding},
    booktitle = {Proceedings of the IEEE/CVF Conference on Computer Vision and Pattern Recognition (CVPR)},
    month     = {June},
    year      = {2023},
    pages     = {21814-21823}
}

@article{PointConT,
title = "Point Cloud Classification Using Content-Based Transformer via Clustering in Feature Space",
journal = "IEEE/CAA Journal of Automatica Sinica",
volume = "11",
number = "JAS-2022-1385",
pages = "231",
year = "2024",
note = "",
issn = "2329-9266",
doi = "10.1109/JAS.2023.123432",
url = "https://www.ieee-jas.net/en/article/doi/10.1109/JAS.2023.123432",
author = {Yahui Liu and Bin Tian and Yisheng Lv and Lingxi Li and Fei-Yue Wang},
}

@article{frank2022so3krates,
  title={So3krates: Equivariant attention for interactions on arbitrary length-scales in molecular systems},
  author={Frank, Thorben and Unke, Oliver and M{\"u}ller, Klaus-Robert},
  journal={Advances in Neural Information Processing Systems},
  volume={35},
  pages={29400--29413},
  year={2022}
}

@article{DFT_reliability,
    author = {Perdew, John P. and Schmidt, Karla},
    title = {Jacob’s ladder of density functional approximations for the exchange-correlation energy},
    journal = {AIP Conference Proceedings},
    volume = {577},
    number = {1},
    pages = {1-20},
    year = {2001},
    month = {07},
    issn = {0094-243X},
    doi = {10.1063/1.1390175},
    url = {https://doi.org/10.1063/1.1390175},
    eprint = {https://pubs.aip.org/aip/acp/article-pdf/577/1/1/12108089/1\_1\_online.pdf},
}

@article{DFT_comp_cost,
author = {Hautier, Geoffroy and Jain, Anubhav and Ong, Shyue},
year = {2012},
month = {05},
pages = {},
title = {From the computer to the laboratory: Materials discovery and design using first-principles calculations},
volume = {47},
journal = {Journal of Materials Science},
doi = {10.1007/s10853-012-6424-0}
}

@article{mat_appl,
  author    = {Ramprasad, Rampi and Batra, Rohit and Pilania, Ghanshyam and Mannodi-Kanakkithodi, Arun and Kim, Chiho},
  title     = {Machine learning in materials informatics: recent applications and prospects},
  journal   = {npj Computational Materials},
  volume    = {3},
  number    = {1},
  pages     = {54},
  year      = {2017},
  doi       = {10.1038/s41524-017-0056-5},
  url       = {https://doi.org/10.1038/s41524-017-0056-5},
  issn      = {2057-3960}
}

@article{solar_cells,
  author    = {Green, Martin A. and Ho-Baillie, Anita and Snaith, Henry J.},
  title     = {The emergence of perovskite solar cells},
  journal   = {Nature Photonics},
  volume    = {8},
  number    = {7},
  pages     = {506--514},
  year      = {2014},
  doi       = {10.1038/nphoton.2014.134},
  url       = {https://doi.org/10.1038/nphoton.2014.134},
  issn      = {1749-4893}
}

@article{superconductor,
  author  = {Bednorz, J. G. and M{\"u}ller, K. A.},
  title   = {Possible high $T_c$ superconductivity in the Ba--La--Cu--O system},
  journal = {Zeitschrift f{\"u}r Physik B Condensed Matter},
  volume  = {64},
  number  = {2},
  pages   = {189--193},
  year    = {1986},
  doi     = {10.1007/BF01303701},
  url     = {https://doi.org/10.1007/BF01303701}
}

@article{batteries,
title = {LixCoO2 (0<x<-1): A new cathode material for batteries of high energy density},
journal = {Materials Research Bulletin},
volume = {15},
number = {6},
pages = {783-789},
year = {1980},
issn = {0025-5408},
doi = {https://doi.org/10.1016/0025-5408(80)90012-4},
url = {https://www.sciencedirect.com/science/article/pii/0025540880900124},
author = {K. Mizushima and P.C. Jones and P.J. Wiseman and J.B. Goodenough},
}

@article{batteries2,
  author    = {Ceder, G. and Chiang, Y.-M. and Sadoway, D. R. and Aydinol, M. K. and Jang, Y.-I. and Huang, B.},
  title     = {Identification of cathode materials for lithium batteries guided by first-principles calculations},
  journal   = {Nature},
  volume    = {392},
  number    = {6677},
  pages     = {694--696},
  year      = {1998},
  doi       = {10.1038/33647},
  url       = {https://doi.org/10.1038/33647},
  issn      = {1476-4687}
}

@article{catalysis,
  author    = {Nørskov, J. K. and Bligaard, T. and Rossmeisl, J. and Christensen, C. H.},
  title     = {Towards the computational design of solid catalysts},
  journal   = {Nature Chemistry},
  volume    = {1},
  number    = {1},
  pages     = {37--46},
  year      = {2009},
  doi       = {10.1038/nchem.121},
  url       = {https://doi.org/10.1038/nchem.121},
  issn      = {1755-4349}
}

@article{MOSELLAMONTORO2021,
title = {2D–3D Geometric Fusion network using Multi-Neighbourhood Graph Convolution for RGB-D indoor scene classification},
journal = {Information Fusion},
volume = {76},
pages = {46-54},
year = {2021},
issn = {1566-2535},
doi = {https://doi.org/10.1016/j.inffus.2021.05.002},
author = {Albert Mosella-Montoro and Javier Ruiz-Hidalgo},
}

@article{DGCNN,
author = {Wang, Yue and Sun, Yongbin and Liu, Ziwei and Sarma, Sanjay E. and Bronstein, Michael M. and Solomon, Justin M.},
title = {Dynamic Graph CNN for Learning on Point Clouds},
year = {2019},
issue_date = {October 2019},
publisher = {Association for Computing Machinery},
address = {New York, NY, USA},
volume = {38},
number = {5},
issn = {0730-0301},
url = {https://doi.org/10.1145/3326362},
doi = {10.1145/3326362},
journal = {ACM Trans. Graph.},
month = oct,
articleno = {146},
numpages = {12}
}

@inproceedings{Mosella2019RAGC,
    author = {Albert Mosella-Montoro and Javier Ruiz-Hidalgo},
    title = {Residual Attention Graph Convolutional Network 
            for Geometric 3D Scene Classification},
    booktitle = {IEEE Conference on Computer Vision Workshop (ICCVW)},
    year = {2019}
}

@inproceedings{pointnet++,
  author       = {Charles Ruizhongtai Qi and
                  Li Yi and
                  Hao Su and
                  Leonidas J. Guibas},
  editor       = {Isabelle Guyon and
                  Ulrike von Luxburg and
                  Samy Bengio and
                  Hanna M. Wallach and
                  Rob Fergus and
                  S. V. N. Vishwanathan and
                  Roman Garnett},
  title        = {PointNet++: Deep Hierarchical Feature Learning on Point Sets in a
                  Metric Space},
  booktitle    = {Advances in Neural Information Processing Systems 30: Annual Conference
                  on Neural Information Processing Systems 2017, December 4-9, 2017,
                  Long Beach, CA, {USA}},
  pages        = {5099--5108},
  year         = {2017},
  url          = {https://proceedings.neurips.cc/paper/2017/hash/d8bf84be3800d12f74d8b05e9b89836f-Abstract.html},
}

@article{Jarvis,
  author    = {Choudhary, Kamal and Garrity, Kevin F. and Reid, Andrew C. E. and DeCost, Brian and Biacchi, Adam J. and Hight Walker, Angela R. and Trautt, Zachary and Hattrick-Simpers, Jason and Kusne, A. Gilad and Centrone, Andrea and Davydov, Albert and Jiang, Jie and Pachter, Ruth and Cheon, Gowoon and Reed, Evan and Agrawal, Ankit and Qian, Xiaofeng and Sharma, Vinit and Zhuang, Houlong and others},
  title     = {The joint automated repository for various integrated simulations (JARVIS) for data-driven materials design},
  journal   = {npj Computational Materials},
  volume    = {6},
  number    = {1},
  pages     = {173},
  year      = {2020},
  doi       = {10.1038/s41524-020-00440-1},
  url       = {https://doi.org/10.1038/s41524-020-00440-1},
  issn      = {2057-3960}
}

@article{opt88,
  title = {Van der Waals density functionals applied to solids},
  author = {Klime\ifmmode \check{s}\else \v{s}\fi{}, Ji\ifmmode \check{r}\else \v{r}\fi{}\'{\i} and Bowler, David R. and Michaelides, Angelos},
  journal = {Phys. Rev. B},
  volume = {83},
  issue = {19},
  pages = {195131},
  numpages = {13},
  year = {2011},
  month = {May},
  publisher = {American Physical Society},
  doi = {10.1103/PhysRevB.83.195131},
  url = {https://link.aps.org/doi/10.1103/PhysRevB.83.195131}
}

@article{MBJ,
  title={A DFT study of BeX (X= S, Se, Te) semiconductor: modified Becke Johnson (mBJ) potential},
  author={Rai, DP and Ghimire, MP and Thapa, RK},
  journal={Semiconductors},
  volume={48},
  pages={1411--1422},
  year={2014},
  publisher={Springer}
}

@article{ensemble_review,
title = {Ensemble deep learning: A review},
journal = {Engineering Applications of Artificial Intelligence},
volume = {115},
pages = {105151},
year = {2022},
issn = {0952-1976},
doi = {https://doi.org/10.1016/j.engappai.2022.105151},
url = {https://www.sciencedirect.com/science/article/pii/S095219762200269X},
author = {M.A. Ganaie and Minghui Hu and A.K. Malik and M. Tanveer and P.N. Suganthan},
}

@article{ensemble_review_2,
title = {A comprehensive review on ensemble deep learning: Opportunities and challenges},
journal = {Journal of King Saud University - Computer and Information Sciences},
volume = {35},
number = {2},
pages = {757-774},
year = {2023},
issn = {1319-1578},
doi = {https://doi.org/10.1016/j.jksuci.2023.01.014},
url = {https://www.sciencedirect.com/science/article/pii/S1319157823000228},
author = {Ammar Mohammed and Rania Kora},
}

@article{multiscale1,
  title={Atomistic-to-continuum coupling},
  author={Luskin, Mitchell and Ortner, Christoph},
  journal={Acta Numerica},
  volume={22},
  pages={397--508},
  year={2013},
  publisher={Cambridge University Press}
}

@article{multiscale22,
  title = {A Vision for the Future of Multiscale Modeling},
  author = {Capone, Matteo and Romanelli, Marco and Castaldo, Davide and Parolin, Giovanni and Bello, Alessandro and Gil, Gabriel and Vanzan, Mirko},
  journal = {ACS Physical Chemistry Au},
  year = {2024},
  volume = {4},
  number = {3},
  pages = {202--225},
  doi = {10.1021/acsphyschemau.3c00080},
  url = {https://doi.org/10.1021/acsphyschemau.3c00080},
  publisher = {American Chemical Society},
}

@article{gga,
  title = {Generalized gradient approximation for the exchange-correlation hole of a many-electron system},
  author = {Perdew, John P. and Burke, Kieron and Wang, Yue},
  journal = {Phys. Rev. B},
  volume = {54},
  issue = {23},
  pages = {16533--16539},
  numpages = {0},
  year = {1996},
  month = {Dec},
  publisher = {American Physical Society},
  doi = {10.1103/PhysRevB.54.16533},
  url = {https://link.aps.org/doi/10.1103/PhysRevB.54.16533}
}


\begin{acknowledgement}

This work was supported by the Spanish Research Agency (AEI) under project PID2024-161868OB-I00, PID2021-122464NB-I00, TED2021-129593B-I00, CNS2023-144561, PID2024-155562NB-I00 and Maria de Maeztu CEX2021-001202-M. E. R. also acknowledges the Generalitat de Catalunya for an ICREA Academia grant.

\end{acknowledgement}

\section{Author contributions statement}

\textbf{À. S.}: Conceptualisation, Methodology, Software, Investigation, Validation, Writing - review \& editing. 
\textbf{A. M-M.}: Conceptualisation, Supervision, Resources, Writing - review \& editing.
\textbf{J. C.}: Conceptualisation, Writing - review \& editing.
\textbf{D. A.}: Conceptualisation, Supervision, Resources,  Writing - review \& editing.
\textbf{S. G-C.}: Conceptualisation, Supervision, Resources, Writing - review \& editing, Funding acquisition.
\textbf{E. R.}: Conceptualisation, Supervision, Resources, Writing - review \& editing, Funding acquisition.
\textbf{J. R-H.}: Conceptualisation, Supervision, Resources, Writing - review \& editing, Funding acquisition.

\section{Conflicts of interest}
There are no conflicts to declare.

\begin{suppinfo}

The supplementary file (PDF) documents the complete PRISM architecture and reproducibility materials: a detailed description of the model components (Superatom encoder; baseline modifications on CartNet; atom and edge encoders; message-passing gates, envelope, and equations), the prediction head, and a scalability/complexity analysis with illustrative schematics, plus a comparison table of per-atom message-passing complexity and parameter counts. It further includes concise dataset summaries for JARVIS 3D DFT (v2021.8.18), Materials Project (2018.6.1), and MatBench tasks (e\_form, jdft2d); training setup (hardware and library versions), optimisation choices, and complete hyperparameter tables for each benchmark. All additional figures, tables, and references cited in the SI are provided within the PDF. 

\end{suppinfo}

\end{document}